\pdfoutput=1
\documentclass[letterpaper, 10 pt, conference]{ieeeconf}  

\IEEEoverridecommandlockouts                              

\usepackage{graphicx}
\graphicspath{{images/}}
\usepackage{amsmath} 
\usepackage{amssymb}  
\usepackage{color}
\usepackage{bm}
\usepackage{siunitx}

\usepackage{hyperref}

\usepackage{bm}

\title{\LARGE \bf
  Transfer learning from synthetic to real images using variational autoencoders for robotic applications
}

\author{Tadanobu Inoue$^{1}$, Subhajit Chaudhury$^{1}$, Giovanni De Magistris$^{1}$ and Sakyasingha Dasgupta$^{2\dagger}$
\thanks{$^{1}$IBM Research - Tokyo, IBM Japan, Japan. \{inouet, subhajit, giovadem\}@jp.ibm.com}
\thanks{$^{2}$LeapMind Inc., Japan. sakya@leapmind.io}
\thanks{$^{\dagger}$ This work was carried out during his position with IBM Research - Tokyo, IBM Japan. }
}

\begin{document}

\maketitle
\thispagestyle{empty}
\pagestyle{empty}

\begin{abstract}

  Robotic learning in simulation environments provides a faster, more scalable, and safer training methodology than learning directly with physical robots.
  Also, synthesizing images in a simulation environment for collecting large-scale image data is easy, whereas capturing camera images in the real world is time consuming and expensive.
  However, learning from only synthetic images may not achieve the desired performance in real environments due to the gap between synthetic and real images.
  We thus propose a method that transfers learned capability of detecting object position from a simulation environment to the real world.
  Our method enables us to use only a very limited dataset of real images while leveraging a large dataset of synthetic images using multiple variational autoencoders.
  It detects object positions 6 to 7 times more precisely than the baseline of directly learning from the dataset of the real images.
  Object position estimation under varying environmental conditions forms one of the underlying requirement for standard robotic manipulation tasks.
  We show that the proposed method performs robustly in different lighting conditions or with other distractor objects present for this requirement.
  Using this detected object position, we transfer pick-and-place or reaching tasks learned in a simulation environment to an actual physical robot without re-training.

\end{abstract}

\section{INTRODUCTION}
\label{sec:intro}
Recent progress in both supervised learning and deep reinforcement learning techniques has provided promising results towards achieving near human-level control in robotics for specific tasks.
In the case of robotic manipulation tasks, large-scale data needs to be collected to take advantage of deep learning techniques, as demonstrated by Levine \textit{et al.}~\cite{LevineISER2016}, who used 14 real robots in parallel for 2 months to collect 800,000 grasp attempts.
Although such methods can perform impressively, their real-world scalability and cost-to-performance ratio are questionable.

On the other hand, robotic learning using physical simulators provides a faster, more scalable, and safer training methodology than learning directly with physical robots.
This has been showcased with impressive performances in simulation driven reinforcement learning~\cite{MnihNIPS2013}\cite{SchulmanICML2015}.
However, equivalent performance in physical robots is still lacking due to the apparent gap in environment domains between simulations and the real world.

Transfer learning with a domain adaptation between simulation environments and the real world~\cite{Rusu}\cite{Higgins}\cite{James} provides promising approaches to overcome this ``reality gap.''
The ability to generalize and adapt between images seen by a robot in simulation environment and those in the real world is crucial for such case of vision-based robotics.

\begin{figure}[thpb]
  \centering
  \includegraphics[width=\linewidth]{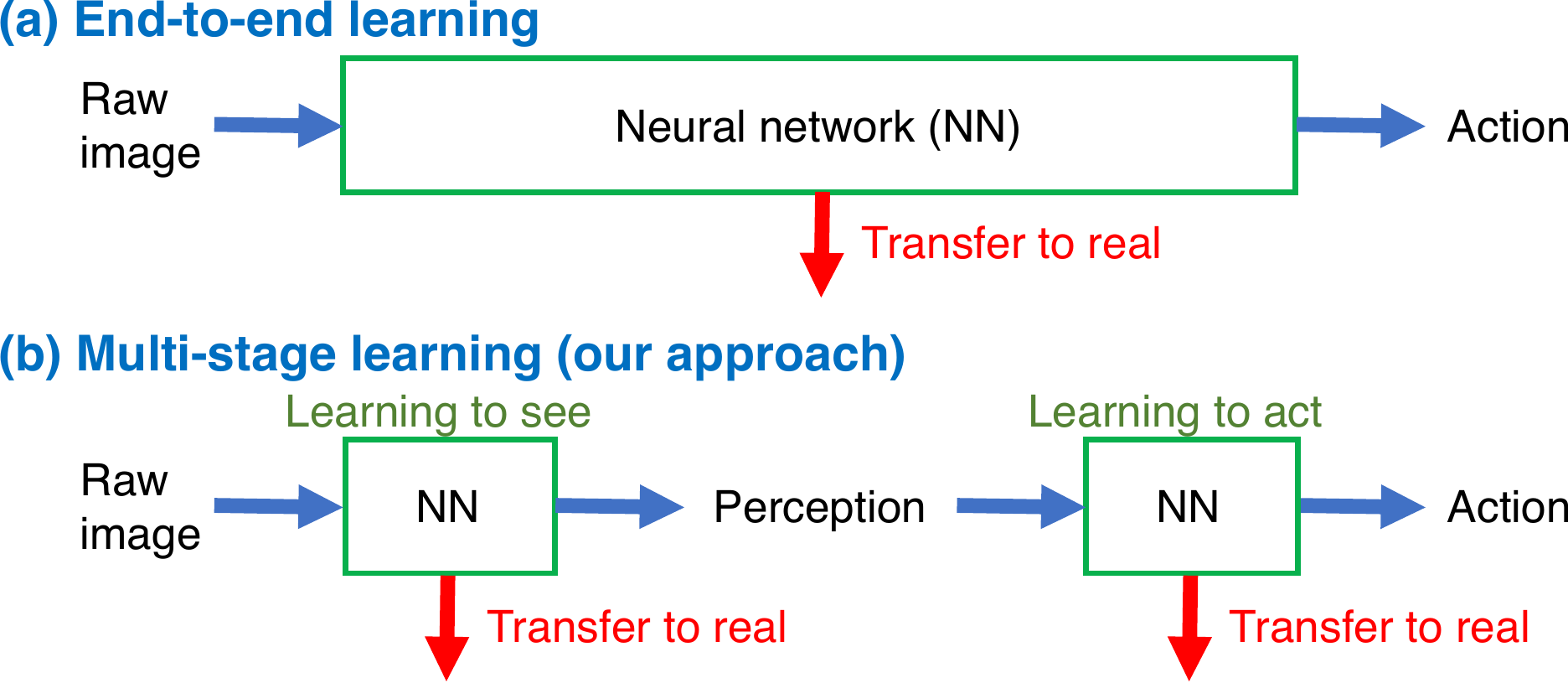}
  \caption{Learning system input-output flow diagrams. (a) End-to-end learning and (b) multi-stage learning in this paper.}
  \label{fig:flow}
\end{figure}

Fig.~\ref{fig:flow} depicts the input-output flow of our approach.
We divide end-to-end learning from image data to action (Fig.~\ref{fig:flow}(a)) into two parts: (1) learning to see and (2) learning to act~\cite{Finn}.
We train these neural network models entirely in simulation environments and then transfer them to the real world (Fig.~\ref{fig:flow}(b)).
Especially, we propose a transfer learning method based on two variational autoencoders (VAE)~\cite{KingmaICLR2014} that generate common pseudo-synthetic images from synthetic and real images for detecting object positions.
Our method successfully overcomes the gap between images synthesized in a simulation and images captured in the real world in order to predict object positions in the real world with less than \SI{1}{\centi\metre} approximation error.
Using this precisely detected object position, we can transfer pick-and-place or reaching tasks learned with a simulated 6-degrees-of-freedom (DOF) robotic arm to an actual physical robot without re-training in the real world.

Learning from synthetic images generally does not enable robots to perform as desired in real environments due to the gap between synthetic and real images.
Shrivastava \textit{et al.}~\cite{Shrivastava} proposed a method to generate realistic images by refining synthesized images to overcome the gap between synthetic and real images.
Santana and Hotz~\cite{Santana} combined a VAE and generative adversarial networks (GANs)~\cite{goodfellow2014generative} to generate realistic road images.
However, these approaches need many unlabeled real images for the adversarial training during the refinement.
Our method enables us to use a very limited dataset of real images, which are typically costly to collect, while leveraging a large dataset of synthetic images that can be easily generated in a simulation environment, for training the neural networks.
Furthermore, it remains invariant to changes in lighting conditions or the presence of other distractor objects.

The rest of paper is organized as follows.
Section \ref{sec:ps} explains the formulation of the transfer learning problem.
Section \ref{sec:method} describes details of our proposed method.
Section \ref{sec:experiments} quantitatively analyzes the method, and Section \ref{sec:robotic_app} presents two robotic applications utilizing our method.
Finally, we conclude the paper in Section \ref{sec:conclusion} with directions for future work.

\section{PROBLEM STATEMENT}
\label{sec:ps}

When we have two labeled image datasets (one synthesized in a simulation environment, and the other captured in the real world), we can assume image instances as $\textbf{X}_S = \{ \bm{x}_S^i\}_{i=1:N} $ and $\textbf{X}_R = \{ \bm{x}_R^i\}_{i=1:M}$ with $S$ representing synthetic image data and $R$ representing real image data, respectively.
Since it is easy to synthesize many images for expected labels in a simulation environment and expensive to capture many images for expected labels in the real world, typically $M << N$.
We thus aim to extract meaningful information, $\bm{y}_R^{(i)} = f(\bm{x}_R^i)$, from the real world images that we can use for subsequent tasks of interest.

However, due to time and cost constraints, it is difficult to collect sufficiently large amounts of real world images to guarantee asymptotic convergence of the function of our interest, i.e. $f$.
We take an approach of modeling a given scene within a simulation environment in order to learn the function mapping, $\bm{y}_S^{(i)} = f(\bm{x}_S^i)$.
This is done based on a large amount of corresponding synthetic images which can be collected easily in simulation.
Given this setting, we want to learn a conditional distribution of synthetic images given the real world images, $p(\bm{x}_S|\bm{x}_R)$, by minimizing the following error,

\begin{equation}
  L = {{}\mathbb{E}}_{p(\bm{x}_S|\bm{x}_R)} ( || f(\bm{x}_R) - f(\bm{x}_S)||^2)
\end{equation}

\noindent
where the expectation with respect to the conditional distribution minimizes the distance between the feature maps obtained from the real images and the feature maps of the corresponding reconstructed synthetic images obtained from the real images based on the conditional distribution $p(\bm{x}_S|\bm{x}_R)$.

\begin{figure}[thpb]
  \centering
  \includegraphics[width=0.8\linewidth]{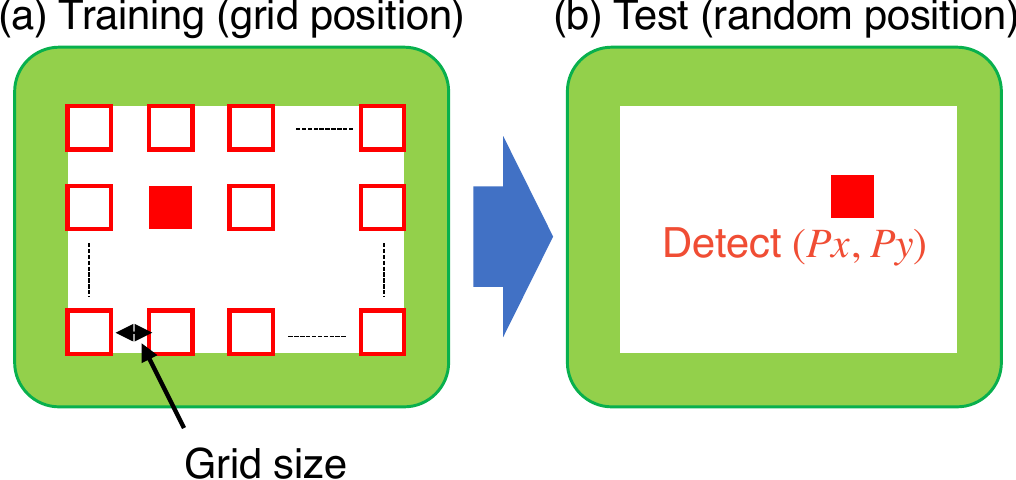}
  \caption{Object position (a) at grid position in training phase and (b) at random position in test phase.}
  \label{fig:cube_pos}
\end{figure}

In this paper, we focus on detecting real object positions from raw RGB-D (red, green, blue, depth) image data using this formulation as our target task for evaluating transfer learning.
First, we train deep neural networks (DNN) with large number of synthetic image data as well as a small number of real image data along with their corresponding object position information.
In order to prepare the data for training, we assume the object is put at a grid position during the training phase to achieve a uniform distribution of object position as shown in Fig.~\ref{fig:cube_pos}(a).
At the time of inference, we perform domain transfer from a synthetic environment to a real one by using the above trained DNN, in order to detect objects placed at random positions.

\section{DETECTING OBJECT POSITIONS USING VAE}
\label{sec:method}

\begin{figure}[thpb]
  \centering
  \includegraphics[width=\linewidth]{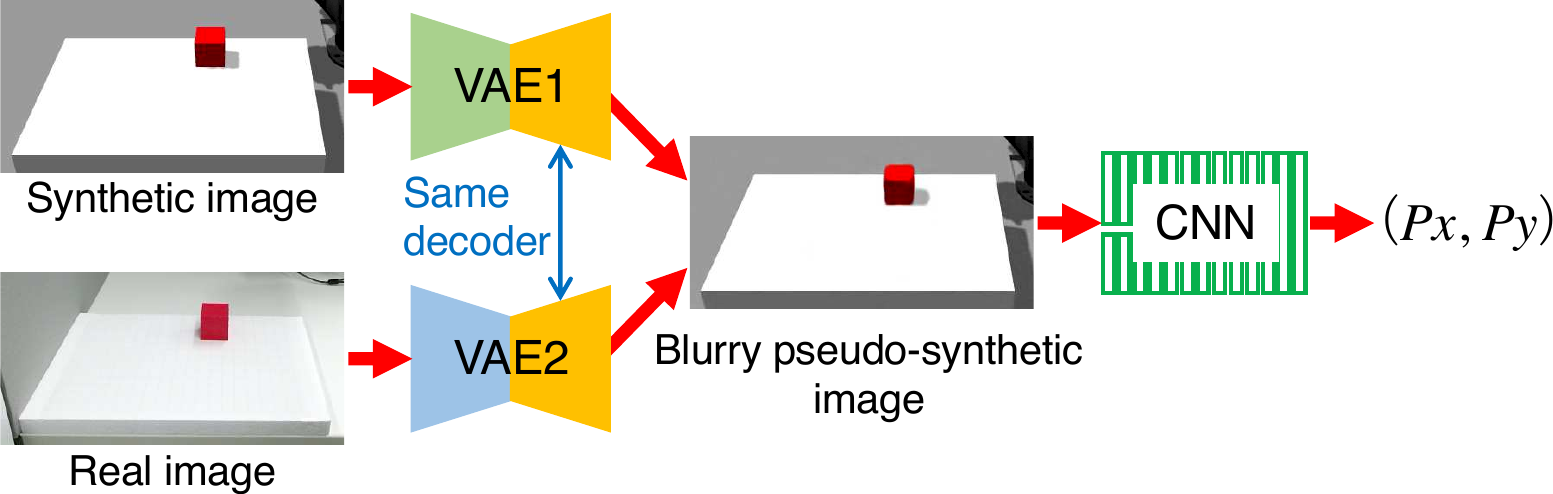}
  \caption{Concept of proposed method for detecting object position.}
  \label{fig:concept}
\end{figure}

Fig.~\ref{fig:concept} depicts the concept of our proposed method.
The core idea of this work is that the distribution of image features may vary between simulated and real environments, but the output labels, like object position, should remain invariant for the same scene.
We use two VAEs for generating similar common images from synthetic and real image data and use this common data distribution to train a convolutional neural network (CNN) to predict the object position with improved accuracy.
Note that although we use two VAEs with distinct encoder layers as generative models for images, they have the same decoder, which is used to train the CNN.
Thus, even if the VAE generates blurry images, the ensuing CNN will learn to predict from this skewed but common image distribution.
Since the CNN can be trained with many generated images from the synthetic domain, we can achieve improved object position estimation from a very limited set of labeled real images.

\begin{figure}[thpb]
  \centering
  \includegraphics[width=0.9\linewidth]{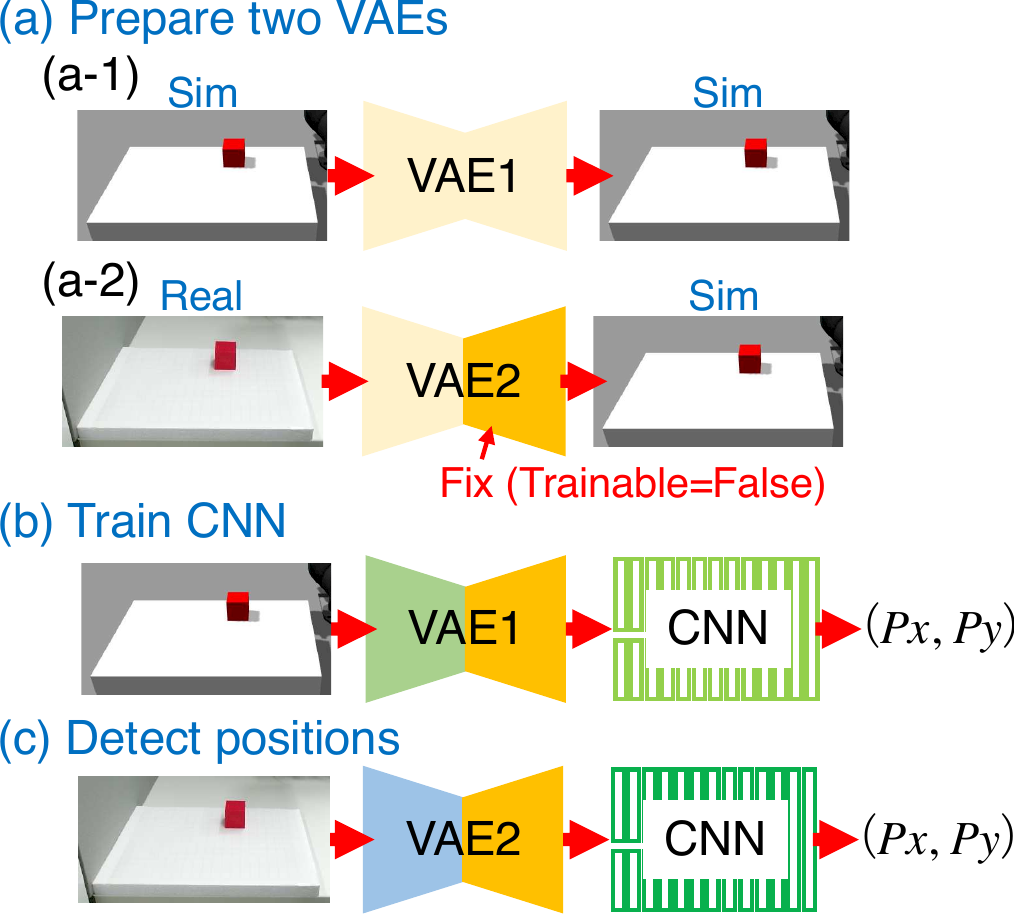}
  \caption{Three steps of our proposed method. (a) Train two VAEs sequentially: (a-1) Traing VAE1 to generate synthetic images. (a-2) Train VAE2 to generate synthetic images from real images. (b) Train CNN to detect object position using VAE1 with synthetic images. (c) Detect real object position using VAE2 and CNN.}
  \label{fig:three_steps}
\end{figure}

Our proposed method consists of three steps as shown in Fig.~\ref{fig:three_steps}:
\begin{itemize}
\item[(a)] Prepare two VAEs that output pseudo-synthetic images from both synthetic and real images.
\item[(b)] Train a CNN to detect object positions using the output of trained VAE in (a).
\item[(c)] Detect object positions by transforming domains using the trained VAE and CNN.
\end{itemize}

First, we prepare two VAEs to generate similar images from synthetic and real images.
We set up a simulation environment that looks similar to the real world and capture large-scale synthetic images$\{\bm{x}_S^i\}_{i=1:N}$ along with corresponding ground-truth object position labels, $\{(t_x^i,t_y^i)\}_{i=1:N}$.
We train VAE1, which encodes and decodes from a synthetic image to the same synthetic image as shown in Fig.~\ref{fig:three_steps}(a-1).

The encoder compresses the input image to the latent representations $\bm{z}$, and the decoder reconstructs the image back from this latent space.
However, using the encoder-decoder results in intractable posterior distribution $p(\bm{z}|\bm{x}_S)$, so we optimize the encoder parameters by variational inference and decoder parameters by minimizing the negative log likelihood of the data.
Using this method, we obtain the optimal parameters $(\bm{\theta}, \bm{\phi})$ by minimizing the lower bound given as,

\begin{equation}
  \scalebox{0.88}{$
    \mathcal{L}_S(\bm{\theta},\bm{\phi};\bm{x}_S^i) = D_{KL}(q_{\bm{\phi}}^S(\bm{z}^i | \bm{x}_S^i) || p_{\theta}(\bm{z}^i)) -
    \mathop{{}\mathbb{E}}(\log p_{\bm{\theta}}(\bm{x}_S^i | \bm{z}))
    $}
\end{equation}

\noindent
where $p_{\theta}(\bm{z}^i)$ is the prior distribution of the latent representation, which is typically the Gaussian with zero mean and unit variance.

We copy the weights of VAE1 to a VAE that has the same structure (VAE2) and then train VAE2, which encodes and decodes from a real image to the corresponding synthetic image as shown in Fig.~\ref{fig:three_steps}(a-2).
During the training, we fix the decoder layers and adapt only the parameters for the encoder part, which receives the real images as input, corresponding to the conditional distribution $q^R_{\bm{\beta}}(\bm{z} | \bm{x}_R)$ with encoder parameters $\bm{\beta}$.
This is equivalent to forcing the latent space obtained from the synthetic and real images to be identical.
Similar work has recently been done by Chaudhury \textit{et al.}~\cite{chaudhury2017conditional}, who explicitly force the latent representation of multi-modal data to be identical during the optimization process.
In this paper, we achieve similar effects for the latent space sharing using only a small dataset of real images by these steps.
We obtain the optimal parameter by minimizing the following lower bound,

\begin{equation}
  \scalebox{0.88}{$
    \mathcal{L}_{R2S}(\bm{\beta};\bm{x}_S^i) = D_{KL}(q^R_{\bm{\beta}}(\bm{z}^i | \bm{x}_R^i) || p_{\theta}(\bm{z}^i)) -
    \mathop{{}\mathbb{E}}(\log p_{\bm{\theta}}(\bm{x}_S^i | \bm{z}))
    $}
\end{equation}

In the above optimization, note that $(\bm{x}_S^i, \bm{x}_R^i)$ are matching pairs of corresponding synthetic and real images.
The learned encoder, $q^R_{\bm{\beta}}(\bm{z} | \bm{x}_R)$, and decoder, $p_{\bm{\theta}}(\bm{x}_S^i | \bm{z})$, can be combined to obtain the desired conditional distribution $p(\bm{x}_S|\bm{x}_R)$, which can generate pseudo-synthetic images as output from the corresponding real image as input.
VAE2 outputs can be subsequently used to obtain accurate object positions from a CNN trained purely in the synthetic image domain.

Next, we train a CNN for detecting object positions as shown in Fig.~\ref{fig:three_steps}(b).
Due to the availability of a large training dataset synthesized in a simulation environment, we can obtain a good prediction by a trained CNN for detecting object positions.
To overcome the gap between synthetic and real images, we use the outputs of the trained VAEs in the above step, instead of using synthetic images directly.

Finally, we can detect object positions in the real world as shown in Fig.~\ref{fig:three_steps}(c).
VAE2 outputs blurry pseudo-synthetic common images, and the CNN trained with the similar common images outputs object position.

There can be alternate strategies in steps (b) and (c).
In the above description, we train the CNN by using VAE1 output, but we can also train a multilayer perceptron (MLP) by using the latent representations obtained from the encoder of VAE1 in step (b).
Since VAE1 and VAE2 have similar latent space structures, we can use latent space output from VAE2 combined with the above trained MLP to detect object position.

\section{EXPERIMENTS}
\label{sec:experiments}

Our proposed method for detecting object positions was evaluated in four experiments:
\renewcommand{\labelenumi}{\Alph{enumi})}
\begin{enumerate}
\item Naive object position detection using a CNN without any generative model. We use this as a baseline.
\item Detection of object positions by the proposed method.
\item Applicability of the proposed method for different shapes.
\item Robustness of the proposed method in different lighting conditions and with other objects present.
\end{enumerate}

\begin{figure}[thpb]
  \centering
  \includegraphics[width=\linewidth]{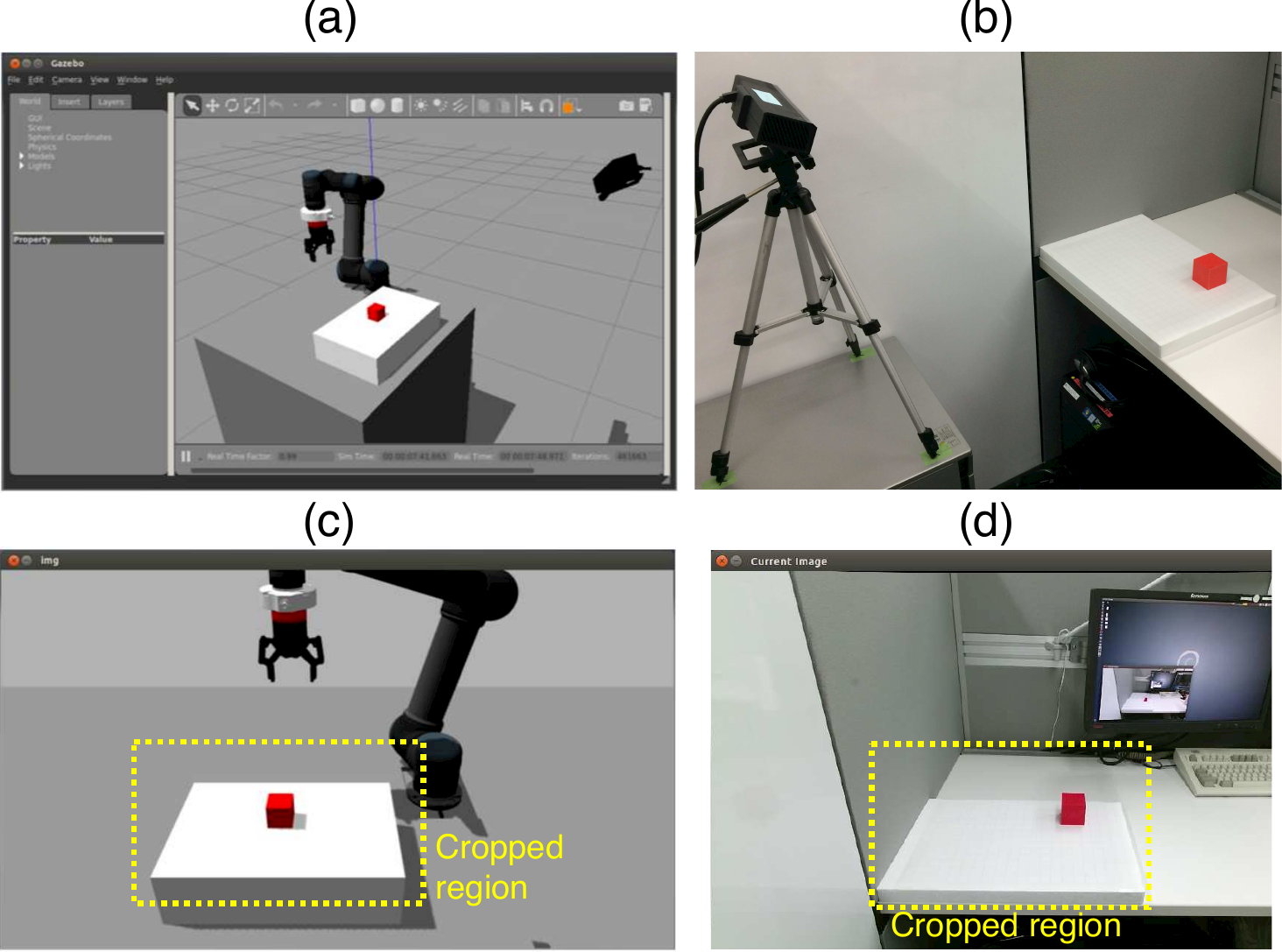}
  \caption{Experimental setup. (a) Capturing synthetic images by simulated Kinect in Gazebo. (b) Capturing real images by physical Kinect in real world. (c) Example of synthetic image captured in (a). (d) Example of real image captured in (b).}
  \label{fig:exp_setup}
\end{figure}

For all experiments, we used a Gazebo\textsuperscript{\textregistered}~\cite{Gazebo} simulation environment and Kinect\textsuperscript{\textregistered}~\cite{Kinect} as shown in Fig.~\ref{fig:exp_setup}.
In Gazebo, an object model is loaded on white Styrofoam ($45\times30\times12$ \SI{}{\centi\metre}) at an intended position as shown in Fig.~\ref{fig:exp_setup}(a).
A corresponding Kinect model is also loaded in Gazebo to capture images of the work space scene.
Fig.~\ref{fig:exp_setup}(c) shows a synthetic image captured by the Kinect in Gazebo.
Additionally, the same object is manually put on the white Styrofoam at an intended position and its image is captured by the physical Kinect as shown in Fig.~\ref{fig:exp_setup}(b).
Fig.~\ref{fig:exp_setup}(d) shows an image captured by the physical Kinect.

The images captured in both Gazebo and the real world were cropped to a smaller region, as shown by the yellow dotted lines in Figs.~\ref{fig:exp_setup}(c) and ~\ref{fig:exp_setup}(d).
The cropped RGB-D image sizes were $400\times200$ pixels for experiments (A)(B) and $472\times280$ for experiments (C)(D).
We used five objects shown in Table~\ref{tb:test_obj} for experiments (A)-(D).

\begin{table}[thbp]
\caption{Objects used in experiments}
\label{tb:test_obj}
\begin{center}
\begin{tabular}{|c|c|c|c|c|c|}
\hline
No. & Color & Shape & Size (\SI{}{\centi\metre}) & Weight ($g$) & Experiment \\
\hline
(1) & red & cube & $5\times5\times5$ & $42$ & (A)(B)(D) \\
\hline
(2) & green & cube & $4\times4\times4$ & $24$ & (C)(D) \\
\hline
(3) & black & cylinder & \shortstack{radius $3.5$ \\ height $1$ } & $17$ & (C)(D)\\
\hline
(4) & blue & \shortstack{triangular \\ prism} & \shortstack{radius $4.5$ \\ height $1$ } & $12$ & (C)(D) \\
\hline
(5) & red & cube & $4\times4\times4$ & $24$ & (D) \\
\hline
\end{tabular}
\end{center}
\end{table}

\begin{figure}[thpb]
  \centering
  \includegraphics[width=\linewidth]{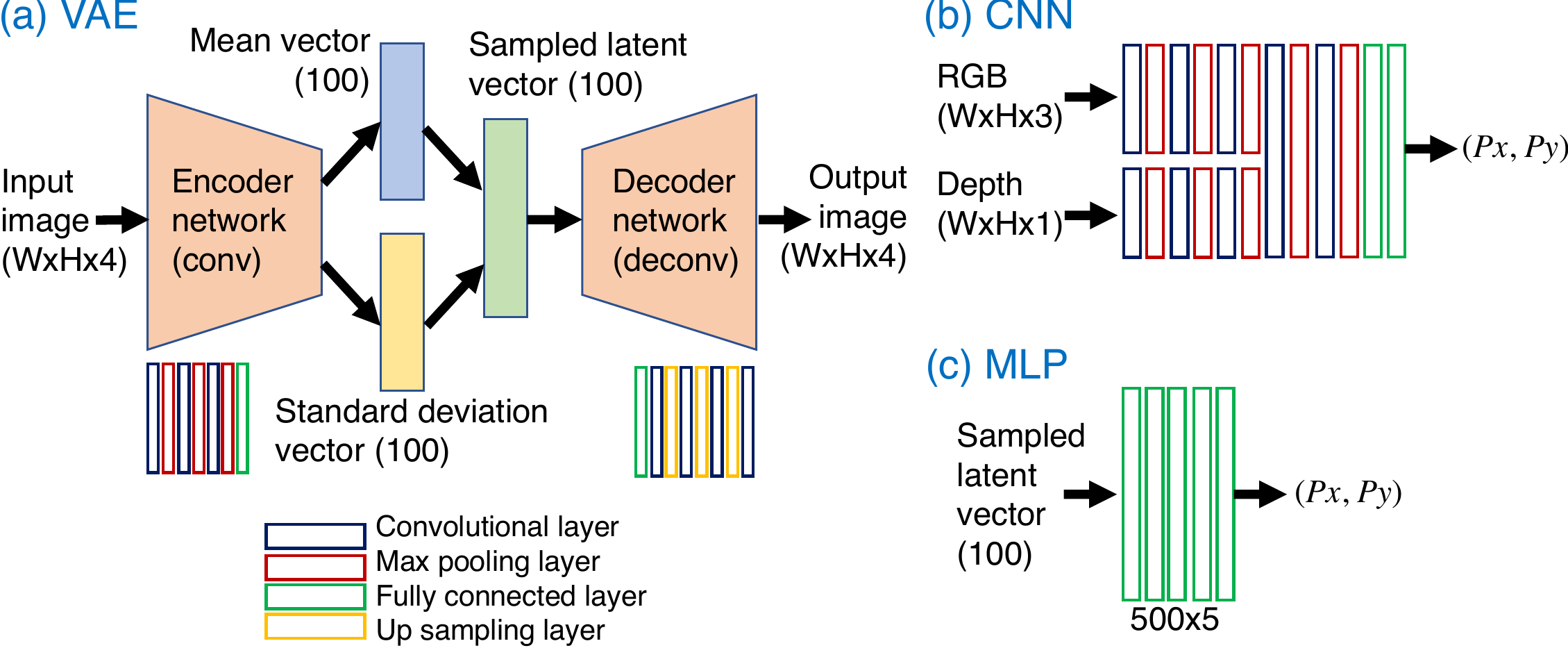}
  \caption{Configuration of neural networks used in experiments. (a) VAE (b) CNN (c) MLP}
  \label{fig:nn}
\end{figure}

For all the experiments, we used the neural network configuration for VAE, CNN and MLP as shown in Fig.~\ref{fig:nn}.

\subsection{Naive object position detection using CNN}

Initially, we detect the red cube (1) position from images captured by Kinect using a CNN.
The red cube is put at an uniform grid position during the training phase and then at a random position during the test phase as shown in Fig.~\ref{fig:cube_pos}.
Making the grid smaller during the training phase enables us to prepare a larger training dataset, where we can expect the cube position to be detected more precisely.
Table~\ref{tb:grid_samples} shows the relationship between the grid size and the number of samples for the red cube on the white Styrofoam.
If we want to obtain \SI{5}{\milli\metre} grid data, we have to capture $4131$ images.
While it is easy to capture $4131$ images in Gazebo, it is highly resource intensive to capture such a large number of images in the real world.
Also, we prepared $200$ synthetic images at random positions for the purpose of evaluation during the later experiments.

\begin{table}[h]
\caption{Grid size and number of samples}
\label{tb:grid_samples}
\begin{center}
\begin{tabular}{|c|c|c|c|c|c|}
\hline
Grid size & \SI{5}{\centi\metre} & \SI{3}{\centi\metre} & \SI{2}{\centi\metre} & \SI{1}{\centi\metre} & \SI{5}{\milli\metre} \\
\hline
Number of samples & $54$ & $150$ & $294$ & $1066$ & $4131$ \\
\hline
\end{tabular}
\end{center}
\end{table}

\begin{figure}[thpb]
  \centering
  \includegraphics[width=\linewidth]{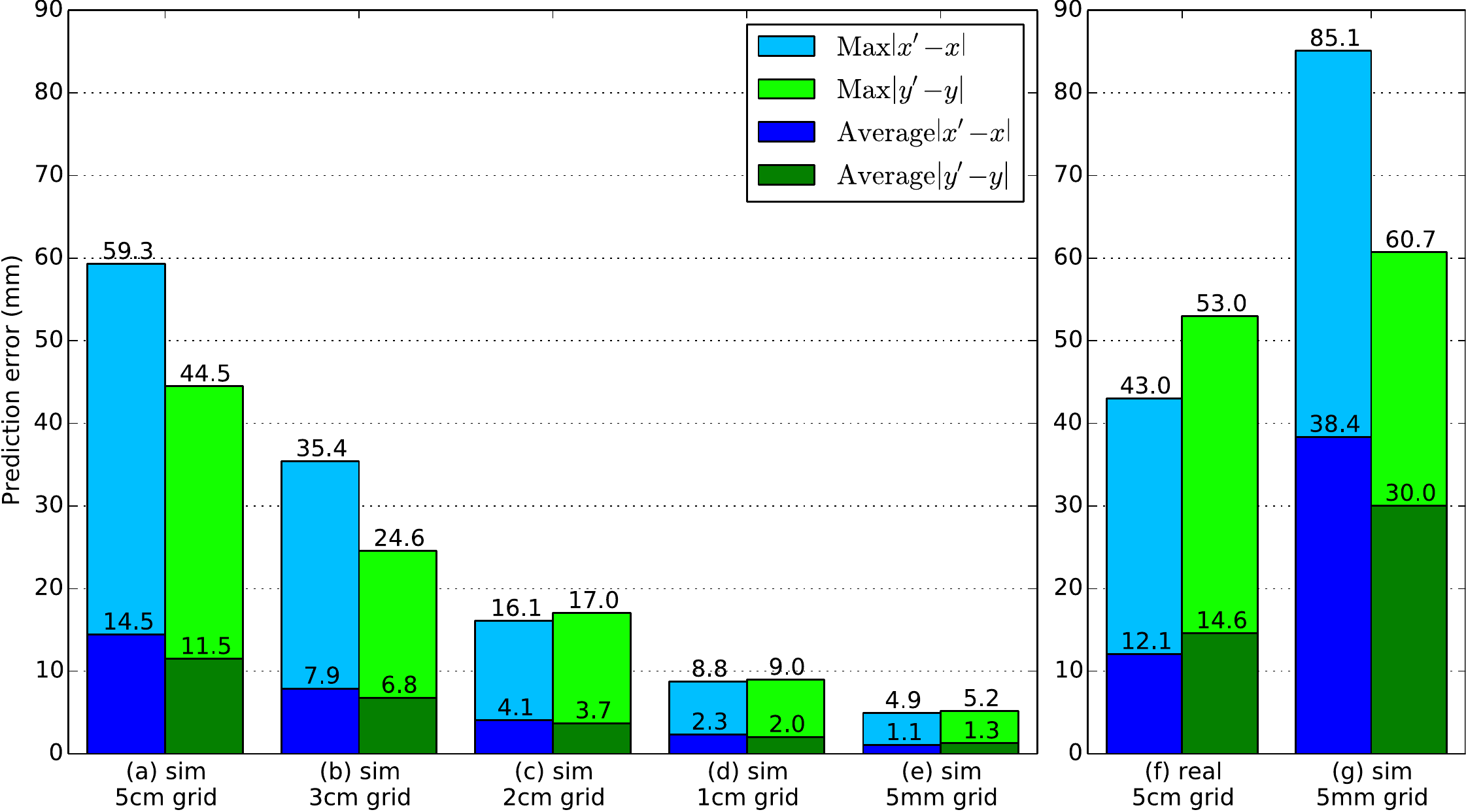}
  \caption{Experimental results of prediction errors. (a)-(e) in Gazebo and (f)-(g) in real world. (a) CNN trained with \SI{5}{\centi\metre} grid, (b) CNN trained with \SI{3}{\centi\metre} grid, (c) CNN trained with \SI{2}{\centi\metre} grid, (d) CNN trained with \SI{1}{\centi\metre} grid, (e) CNN trained with \SI{5}{\milli\metre} grid, (f) CNN trained with \SI{5}{\centi\metre} grid real data, and (g) CNN trained with \SI{5}{\milli\metre} grid synthetic data}
  \label{fig:pred_baseline}
\end{figure}

Fig.~\ref{fig:pred_baseline} shows the prediction errors: the difference between the true position and position detected by the trained CNN.
The dark blue and dark green bars show the average position estimation error values for $x$ and $y$ coordinates, respectively.
The light blue and light green bars show the maximum value of error for $x$ and $y$ coordinates, respectively.
In Gazebo, the errors can be reduced by using a more fine-pitch grid during the training phase as shown in Figs.~\ref{fig:pred_baseline}(a)-(e).
When we use \SI{5}{\milli\metre} grid synthetic data for the training, we can obtain about \SI{1}{\milli\metre} errors on average and \SI{5}{\milli\metre} errors at worst in Gazebo.

In order to detect real object positions, we captured $54$ real images at \SI{5}{\centi\metre} grid positions for the training data.
Moreover, we prepared $50$ real images at random positions for the later evaluation.
If we can obtain cube images at more fine-pitch grid positions in the real world as well as in Gazebo, we can expect better precision.
However, in practice it is very challenging to capture numerous images in the real world.
The bar charts in Fig.~\ref{fig:pred_baseline}(f) show the prediction errors by the CNN trained with the $54$ real images.
As observed, these results are similar to those for the CNN trained with $54$ synthetic images in Gazebo.
We set this result as the baseline for the later comparison.

We tried to feed the real test images to the CNN trained with $4131$ images synthesized in Gazebo, and the corresponding prediction errors are shown as the bar charts in Fig.~\ref{fig:pred_baseline}(g).
As expected, the result for this naive usage of a model trained in Gazebo is worse than the baseline result in Fig.~\ref{fig:pred_baseline}(f).
This is primarily due to the ``reality gap'' between the synthetic and the real images.

\subsection{Detection of object positions by proposed method}

We applied our proposed method for the same red cube (1) used in experiment A.
We used $4131$ synthetic images at \SI{5}{\milli\metre} grid positions for training VAE1 and $54$ real images and $54$ synthetic images at \SI{5}{\centi\metre} grid positions for training VAE2.
Note that no data augmentation of these image data were performed~\cite{Bateux}.

\begin{figure}[thpb]
  \centering
  \includegraphics[width=\linewidth]{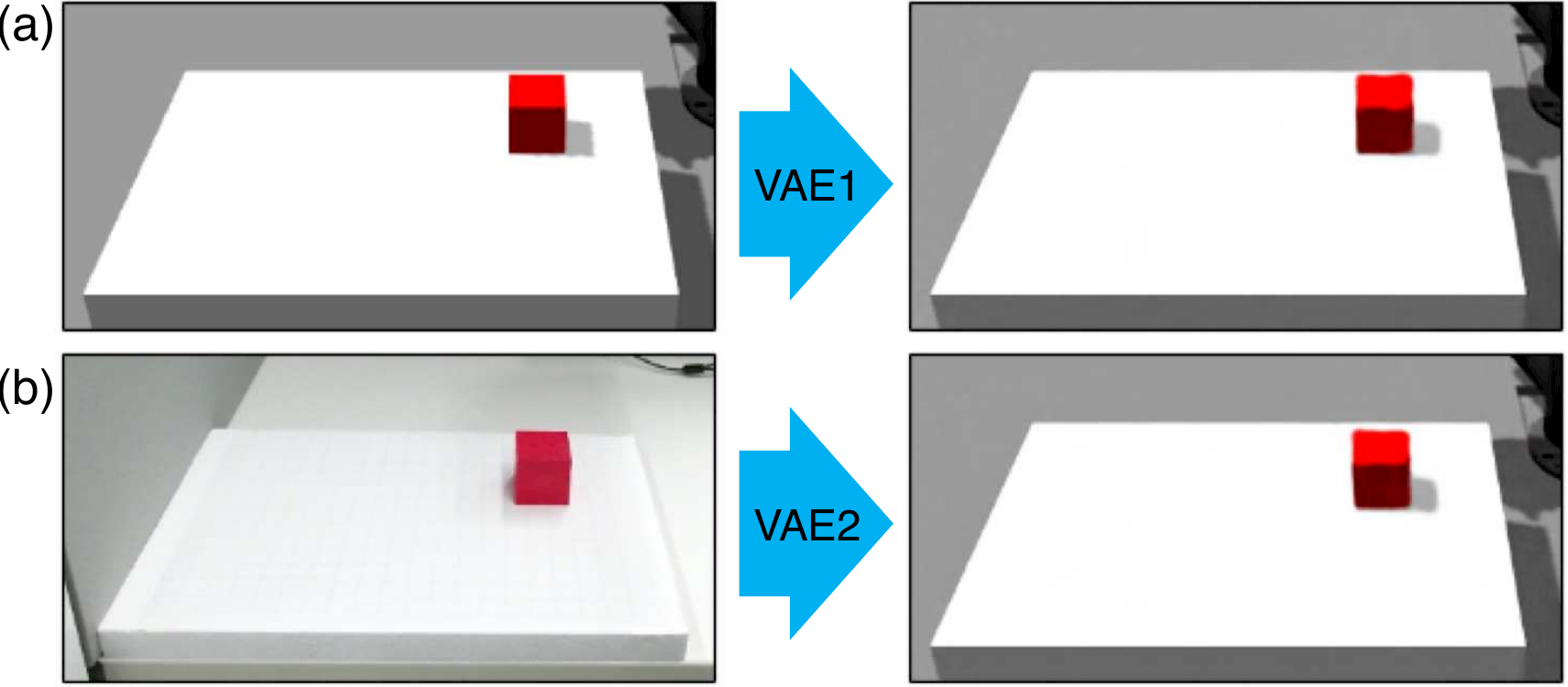}
  \caption{Images generated by two VAEs. (a) VAE1 output from synthetic image input. (b) VAE2 output from real image input.}
  \label{fig:vae_outputs_red}
\end{figure}

Fig.~\ref{fig:vae_outputs_red}(a) shows VAE1 output from a synthetic image input, and Fig.~\ref{fig:vae_outputs_red}(b) shows VAE2 output from a real image input.
Fig.~\ref{fig:vae_outputs_red} shows that we can obtain similar images that look like blurry synthetic images by using these two VAEs as we expected.
The mean squared error (MSE) was $9.9 \times 10^3$ on average for two input images and $2.7 \times 10^{-3}$ on average for two output images.
This serves as a quantitative metric to show that our proposed VAEs improve the similarity between output images from real and synthetic domains.
Although the white Styrofoam is put on a white desk and there is a wall on the left-hand side in the real world image, the white Styrofoam in the VAE output is put on a gray desk as shown in Fig.~\ref{fig:vae_outputs_red}(b).
VAE2's output does not contain the real world wall but does contain a lower part of a robot arm and a similar shadow of the red cube.

\begin{figure}[thpb]
  \centering
  \includegraphics[width=\linewidth]{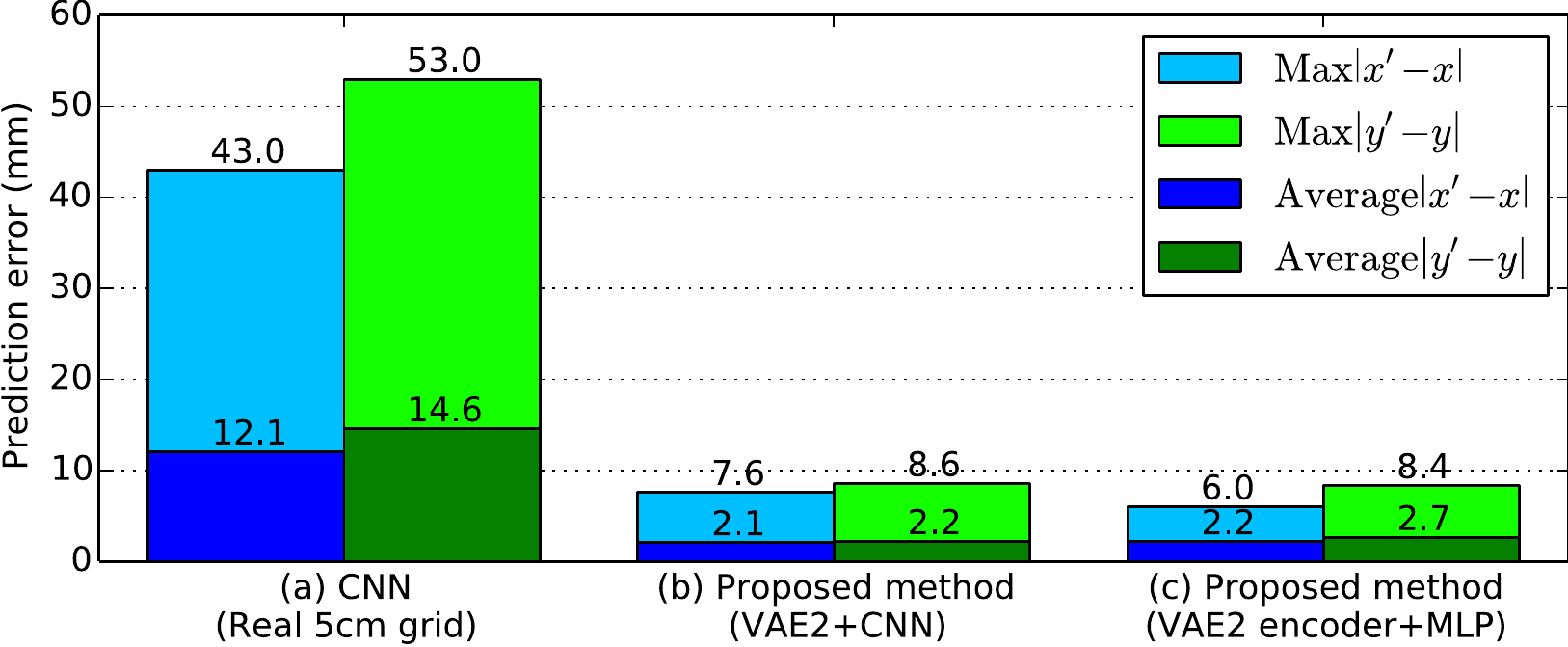}
  \caption{Predicted real object position errors. (a) Baseline method. (b) Proposed method of VAE2 and CNN option. (c) Proposed method of VAE2 encoder and MLP option.}
  \label{fig:precision_vae_latent}
\end{figure}

After successfully learning to generate images with the VAE, we train the CNN/MLP for detecting cube positions using $4131$ synthetic images generated in Gazebo.
Fig.~\ref{fig:precision_vae_latent} shows the prediction errors of the red cube (1) in the real world.

As observed in Fig.~\ref{fig:precision_vae_latent}(b), our proposed method of using the generated images of VAE2 to train a CNN results in a considerable reduction (nearly $6-7$ times) of the prediction error, as compared to the the baseline results of Fig.~\ref{fig:precision_vae_latent}(a).
Furthermore as observed by the results in Fig.~\ref{fig:precision_vae_latent}(c), the alternative method of using the output of VAE2 encoder to train a MLP also results in similar good performance.
These results suggest that the performance benefit clearly stems from the domain adaptation obtained by our VAE training method and it does not depend much on the choice of CNN or MLP.

\subsection{Applicability of proposed method for different shapes}

\begin{figure}[thpb]
  \centering
  \includegraphics[width=\linewidth]{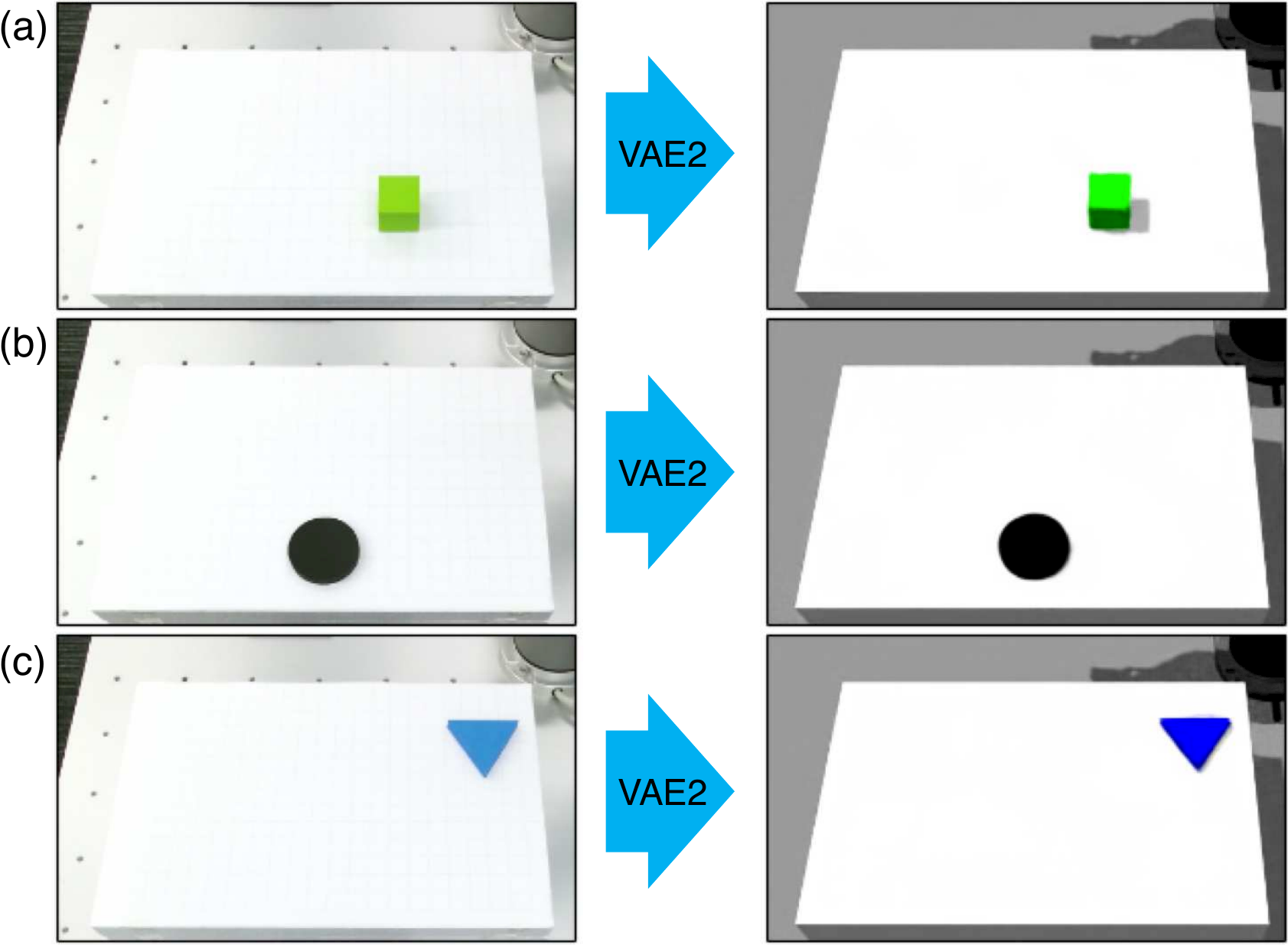}
  \caption{Images generated by VAE2. (a) Green cube. (b) Black cylinder. (c) Blue triangular prism.}
  \label{fig:vae_outputs_green_black_blue}
\end{figure}

We applied our proposed method to other shapes and colors as well.
At this time, we put the white Styrofoam on a table in front of a real robot arm.
For these experiments, we used three of the objects in Table~\ref{tb:test_obj}: green cube (2), black cylinder (3), and blue triangular prism (4).
We prepared two VAEs in our proposed method for each object, and the corresponding outputs of VAE2 are shown in Fig.~\ref{fig:vae_outputs_green_black_blue}.
In all three cases, VAE2 learns to successfully generate pseudo-synthetic images from the corresponding real images.

\subsection{Robustness of proposed method against different lighting conditions and distractor objects}

\begin{figure}[thpb]
  \centering
  \includegraphics[width=\linewidth]{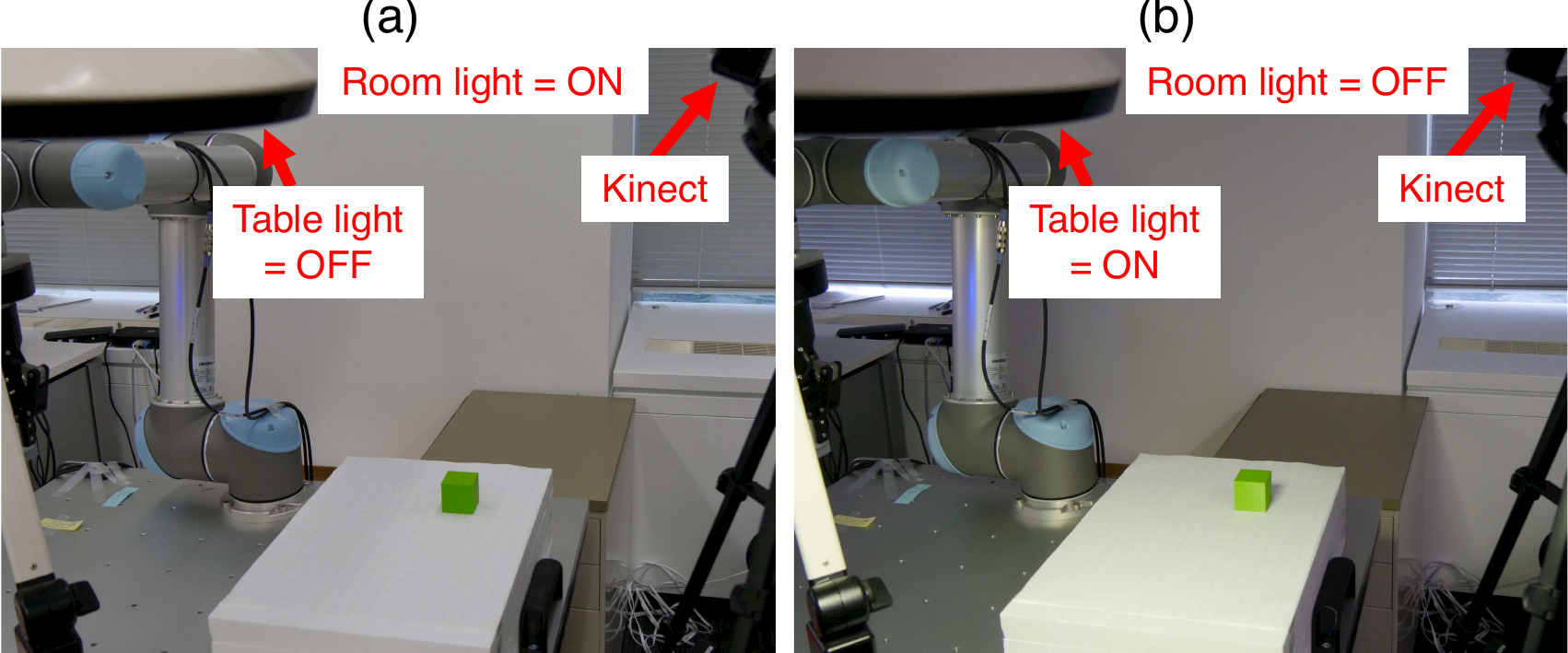}
  \caption{Object position detection under different lighting conditions. (a) Room light turned on and table light turned off. (b) Room light turned off and table light turned on.}
  \label{fig:lighting_conditions}
\end{figure}

In addition, we assessed the robustness of the proposed method in different lighting conditions and with other objects present.
We usually kept the room light of our experimental space turned on as shown in Fig.~\ref{fig:lighting_conditions}(a).
We turned off the room light and turned on a table light instead for a different lighting condition as shown in Fig.~\ref{fig:lighting_conditions}(b)

\begin{figure}[thpb]
  \centering
  \includegraphics[width=\linewidth]{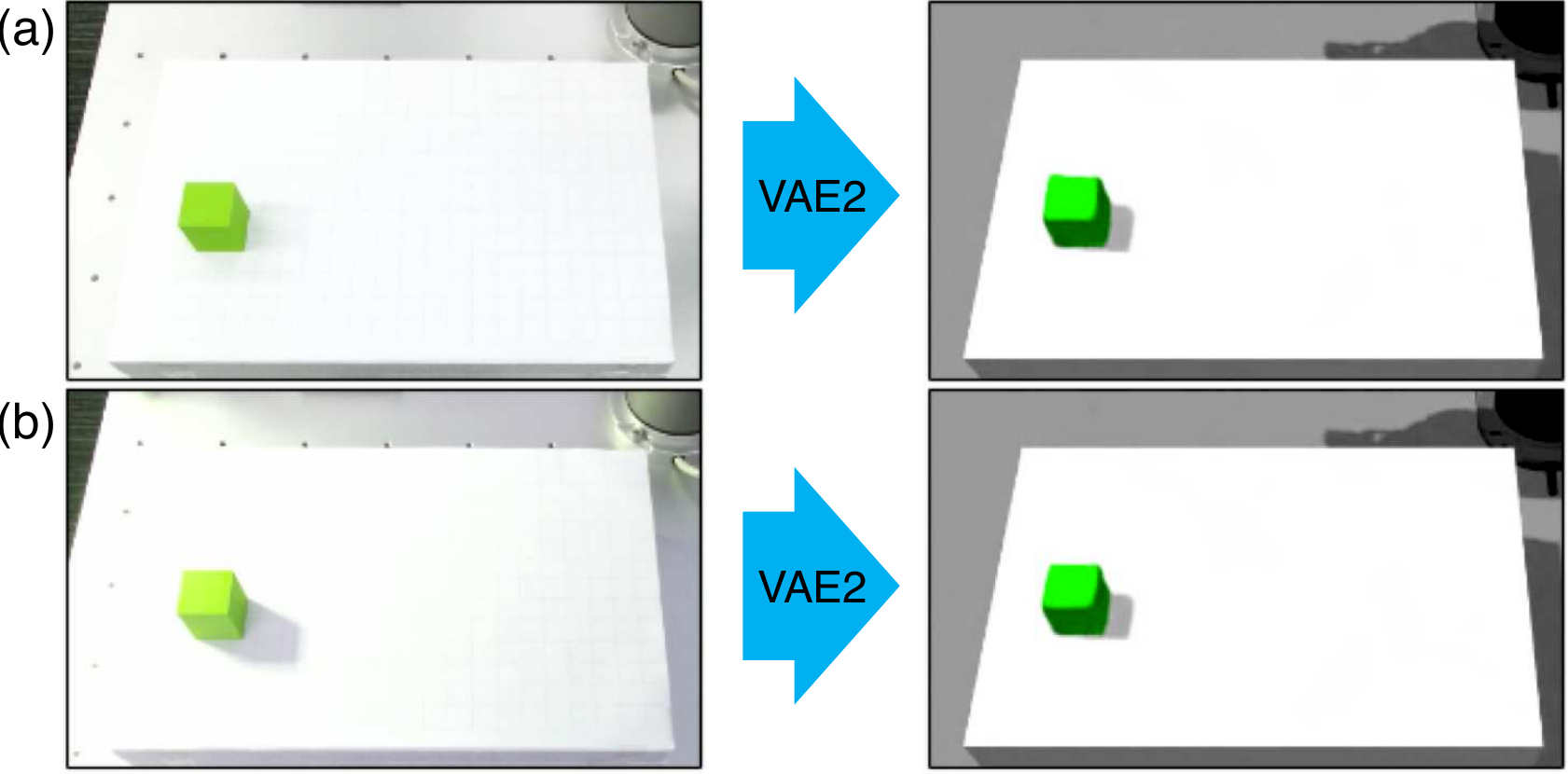}
  \caption{Images generated by VAE2 under different lighting conditions. (a) Room light turned on and table light turned off. (b) Room light turned off and table light turned on}
  \label{fig:vae_outputs_lighting}
\end{figure}

Fig.~\ref{fig:vae_outputs_lighting} shows the VAE2 outputs under two lighting conditions.
VAE2 is trained with captured images under the default lighting condition shown in Fig.~\ref{fig:lighting_conditions}(a) and applied to the captured images under the other lighting condition without any re-training.
The images on the left-hand side are raw images captured by the physical Kinect.
Although the lighting condition in Fig.~\ref{fig:lighting_conditions}(b) is actually quite darker, the brightness levels of captured images are made similar due to Kinect's auto-brightness functionality.
The green color of the object is much lighter and the shadow is cast longer to the right in Fig.~\ref{fig:vae_outputs_lighting}(b) than in the default lighting condition.
As observed from the right side of  Figs.~\ref{fig:vae_outputs_lighting}(a) and ~\ref{fig:vae_outputs_lighting}(b), in both cases the VAE2 learns to generate very similar pseudo-synthetic images.
Although the MSE between the real images under different lighting conditions was $6.9 \times 10^2$, the generated images result in a much smaller MSE of $5.8 \times 10^{-4}$ on average.
This shows that using our proposed method we can effectively deal with differences in the scene caused by varying lighting conditions.

\begin{figure}[thpb]
  \centering
  \includegraphics[width=\linewidth]{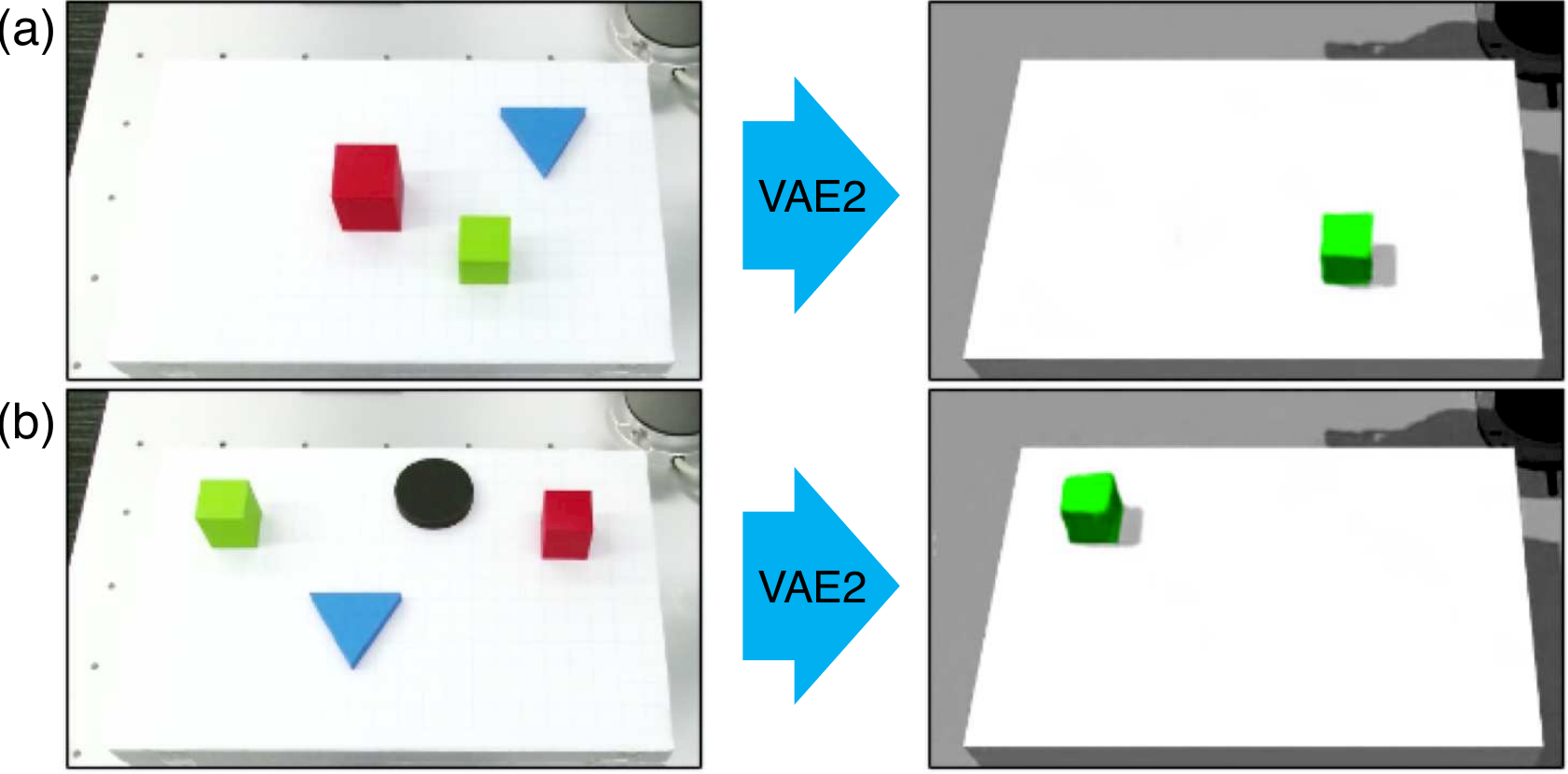}
  \caption{Images generated by VAE2 with presence of distractor objects. (a) Scene containing green cube (2), red cube (1), and blue triangular prism (4). (b) Scene containing green cube (2), red cube (5), black cylinder (3) and blue triangular prism (4). In both cases the target object is the green cube (2)}
  \label{fig:vae_outputs_multiple}
\end{figure}

In the second set of experiments, we evaluated the robustness of the proposed method against the presence of multiple distractor objects in the same scene.
Fig.~\ref{fig:vae_outputs_multiple} shows the generated output images from VAE2 when fed with real images containing multiple other objects.
In this case, the VAE2 was trained with a single green cube object and it was subjected to the newly captured images with multiple objects without any further re-training.
As shown in the right side of Fig.~\ref{fig:vae_outputs_multiple}, the VAE2 continues to robustly generate pseudo-synthetic images with only the green cube while completely ignoring the other objects in the same scene.
Therefore, this selectivity is quite useful for detecting the position of target objects even in the presence of numerous distractor objects of varying colors and shapes.

\begin{figure}[thpb]
  \centering
  \includegraphics[width=\linewidth]{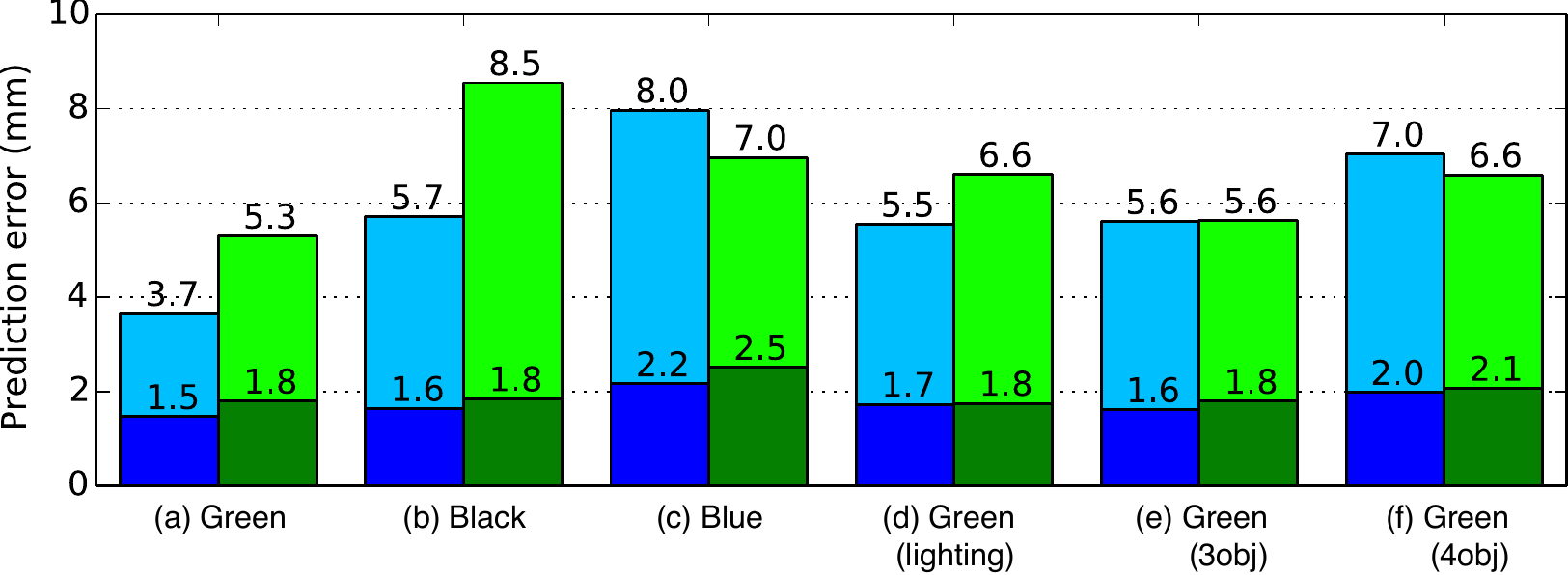}
  \caption{Experimental results for prediction errors. (a) Green cube, (b) black cylinder, (c) blue triangular prism, (d) green cube under different lighting condition, (e) green cube with two other objects, and (f) green cube with three other objects}
  \label{fig:precision_various}
\end{figure}

Fig.~\ref{fig:precision_various} shows the experimental results for prediction errors for the above cases.
We can see our proposed method can be successfully applied to differently shaped objects (Figs.~\ref{fig:precision_various}(a)-(c)) and perform robustly in different lighting conditions and with other objects present (Figs.~\ref{fig:precision_various}(d)-(f)).

\section{ROBOTIC APPLICATIONS}
\label{sec:robotic_app}

In this section, we present the results of robotic tasks using the object position obtained by the proposed transfer learning method.
Specifically, we perform two robot arm manipulations: (a) a pick-and-place task using finite state machines (FSM) trained in a supervised learning manner and (b) a reaching task using model-free control trained by reinforcement learning.
The position of the desired object was estimated using our proposed VAE encoder with MLP method in both cases.
Furthermore, in both cases the entire training were performed with a simulated robot arm and the learned policies were directly transferred to the real robot, generating trajectories in the real world without additional retraining.

We used Gazebo for the simulation environment and Kinect for the RGB-D camera.
We used a UR5\textsuperscript{\textregistered} robot arm from Universal Robots\textsuperscript{\textregistered} and a two-finger gripper from Robotiq\textsuperscript{\textregistered} for robotic manipulation in the real world.
The robot arm consists of 6 joints, and the gripper can be controlled with an analog value between $0.0$ (open) and $1.0$ (close).
In Gazebo, we used a corresponding simulated UR5 with simulated 2-finger gripper and it was controlled with a position based controller.

\subsection{Pick and Place Task Using Learned FSM}

Rahmatizadeh \textit{et al.}~\cite{Rahmatizadeh} propose an approach where the user demonstrates a manipulation task in a virtual environment.
They use the collected demonstrations to train a long short-term memory (LSTM) neural network to control the robots movements.
When applied to the real world, Rahmatizadeh \textit{et al.} use objects annotated with 2D QR code markers to detect their positions.
However, here we applied our proposed method to detect object positions directly from RGB-D images captured by Kinect.

We defined a pick-and-place task as transitions of an FSM that a green cube is picked up and put on a red cube.
The same as De Magistris \textit{et al.}~\cite{Magistris}, we executed a program for the transitions of the FSM repeatedly and recorded six robot arm joint angles $(q_0^{t=T}, q_1^{t=T}, ... q_5^{t=T})$ and one gripper position $g^{t=T}$ at in each state and the initial positions of red $(Pr_x^{t=0}, Pr_y^{t=0})$ and green $(Pg_x^{t=0}, Pg_y^{t=0})$ cubes obtained from Gazebo for supervised training data.
The training dataset was collected from over $7000$ random cubes positions in an accelerated simulation environment with a real-time factor of $6 - 8$.

\begin{figure}[thpb]
  \centering
  \includegraphics[width=\linewidth]{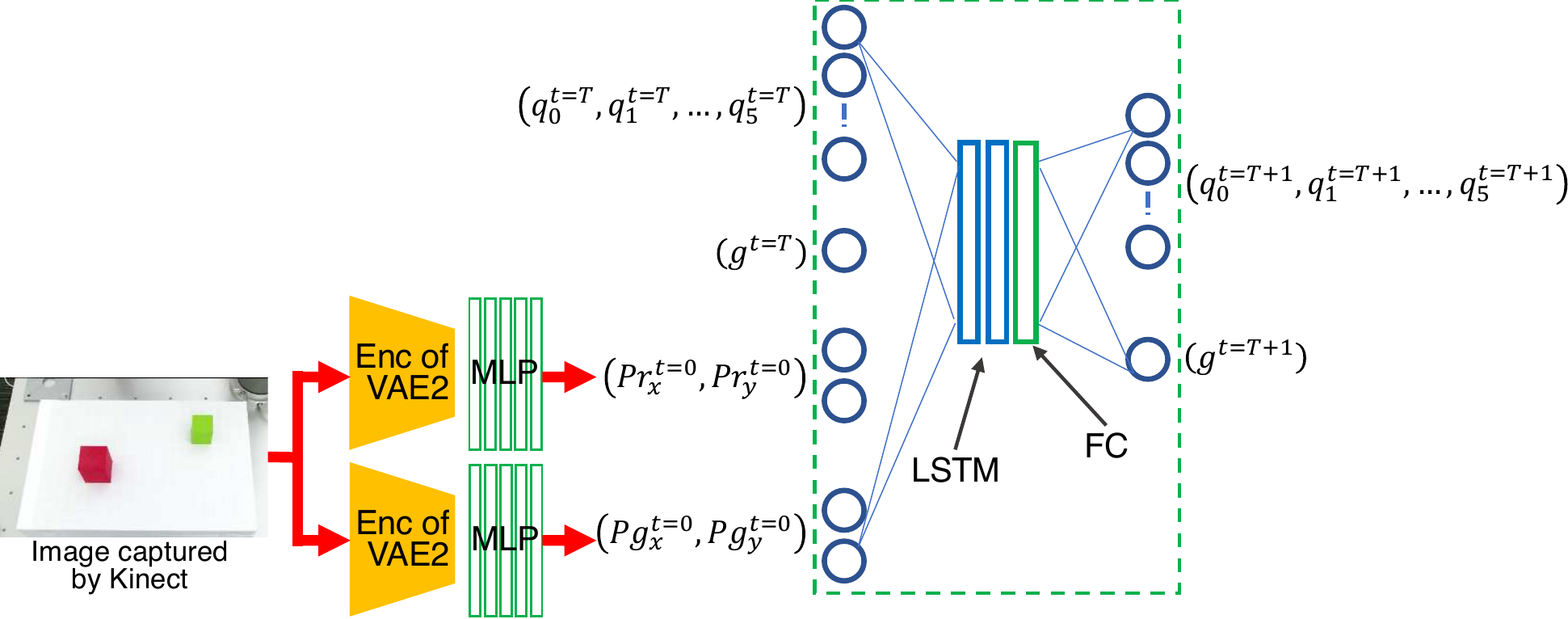}
  \caption{LSTM neural network combined with proposed object position detection method for pick and place task}
  \label{fig:pick_and_place_diagram}
\end{figure}

Fig.~\ref{fig:pick_and_place_diagram} shows the diagram of the LSTM neural network fed with the estimated object positions directly from real image data based on our VAE-MLP pipeline.
The LSTM neural network has 11 inputs (six robot arm joint angles, one gripper position at the current state, and four initial positions of two cubes) and seven outputs (six robot arm joint angles and one gripper position at the next state).

We trained the LSTM based robot controller by using these supervised data and then it could complete the sequence of the task in Gazebo with a $100\%$ success rate in $10$ trials.
Finally the learned policy was transferred to the real robot without any retraining.
We obtain the estimated positions of the red and green cubes based on our proposed method and fed these position data into the LSTM that was trained entirely in Gazebo.
This network was then able to predict the sequence of actions needed for the task in the real world with $100\%$ success rate, across 10 trials.
The successful experimental results for the pick and place task are shown in the video (see \url{https://youtu.be/Wd-1WU8emkw}).

\subsection{Reaching Task Using Model-Free Deep Deterministic Policy Gradient Learning}

To train the robotic arm for reaching a 2D point on the table top, we performed model-free training of the simulated robotic arm in Gazebo.
Angular position, angular velocity, and target $(x, y, z)$ position were used as state information for the robotic arm.
During training, the target $(x, y)$ position was set randomly on the table top, whereas for testing, the target $(x, y)$ position was obtained by the proposed method.
The height $z$ was fixed as the table top height.
Continuous angular acceleration values for each joint were used as six-dimensional action space.

In order to do reinforcement learning with continuous actions, we used the deep deterministic policy gradient (DDPG) algorithm~\cite{Lillicrap} that learns to compute continuous actions from state information input at each time step.
We used the following reward function at each time-step,
\begin{equation}
  r_t = k_1\left\Vert\left( \bm{x}_{\rm{cur}} - \bm{x}_{\rm{target}} \right) \right\Vert^2 + k_2\left\Vert\left( \bm{\theta}_{\rm{cur}} - \bm{\theta}_{\rm{target}} \right) \right\Vert^2
\end{equation}

\noindent
where $\bm{x}_{\rm{cur}}$ and $\bm{x}_{\rm{target}}$ are the current position of the robotic arm end-effector and the desired position of the object to be reached.
The terms $\bm{\theta}_{\rm{cur}}$ and $\bm{\theta}_{\rm{target}}$ represent the roll, pitch, and yaw angles of the current and desired pose of reaching, respectively.
The reward function was designed so as to ensure that the robotic arm reaches the target position at a viable pose to enable grasping.
The first term in the reward function minimizes the Euclidean distance between current 3D location of the robot's end-effector and target 3D location.
The second term minimizes the $L2$ difference between the Euler angles of the current and desired poses as a soft constraint.
We annealed the value of $k_2$ from a high to low value during the training period to ensure that the agent learns to reach with an accurate pose during early phases of training.
Similarly, $k_1$ was varied from a low to high value to ensure accurate reaching of the end-effector to the object.

The learned policy for reaching a point on the table top using the DDPG algorithm in Gazebo was directly transferred to the real robot.
The real environment was set up, and the object location on the table was computed using the proposed method from real images captured by Kinect.

After training, the learned policy for reaching an object on the table top using DDPG in Gazebo was directly transferred to the real robot, without any further reinforcement learning with the physical robot arm.
Finally, in the real environment the object location on the table was computed directly from real raw RGB-D images captured by Kinect using our VAE-MLP pipeline.
The successful experimental results for the object reaching task are shown in the video (see \url{https://youtu.be/Wd-1WU8emkw}).

\section{CONCLUSION AND FUTURE WORK}
\label{sec:conclusion}
We present a transfer learning method for transferring the ability of detecting object positions learned in a simulation environment to the real world.
The proposed method can detect object positions more precisely than the baseline method of training a CNN or MLP directly from real image data.
Our method performs robustly in different lighting conditions and with other objects present.

In addition, we applied the proposed method to robotic applications with robot controllers learned in the simulation environment and executed them in the real world combined with our proposed object position detection method.
We show that a robotic arm can complete both a pick-and-place task and a reaching task in a real environment without any additional training.
Furthermore it works independent of the type of learning mechanism i.e. supervised or reinforcement learning.

In this paper, we use only simple shapes with fixed poses and different colors for multiple object cases.
For the next step, we plan to enhance our method for various type of shapes with varied poses and demonstrate object selectivity among same color objects.
Transfer learning on complex natural images with arbitrary camera orientation is also a prominent direction for future research.
In addition, we will tackle transfer learning of more complex robotic applications like contact based precise assembly tasks~\cite{Inoue}.



\end{document}